\documentclass[11pt,a4paper]{article}
\usepackage[hyperref]{emnlp-ijcnlp-2019}

\usepackage{times}
\usepackage{CJKutf8}
\usepackage{pbox} 
\usepackage{booktabs}
\usepackage{graphicx}
\usepackage{xcolor}

\usepackage{arabtex}
\usepackage[utf8]{inputenc}
\usepackage[LFE,LAE,T1]{fontenc}

\usepackage{expl3}
\usepackage{l3regex}
\ExplSyntaxOn
\tl_new:N \l_substituteNumbers_tl
\cs_new:Npn \substituteNumbers #1 {
    \tl_set:Nn \l_substituteNumbers_tl {#1}
    \regex_replace_all:nnN {[0-9]+} {  \c{LR}\cB {\0 \cE}  } \l_substituteNumbers_tl
    \tl_use:N \l_substituteNumbers_tl
}
\ExplSyntaxOff

\renewenvironment{abstract}%
         {\centerline{\large\bf Abstract}%
          \begin{list}{}%
             {\setlength{\rightmargin}{0.6cm}%
              \setlength{\leftmargin}{0.6cm}}%
           \item[]\ignorespaces}%
         {\unskip\end{list}}

\aclfinalcopy % Uncomment this line for the final submission
\title{Handling Syntactic Divergence in \\
Low-resource Machine Translation}

\author{Chunting Zhou, Xuezhe Ma, Junjie Hu, Graham Neubig \\
  Language Technologies Institute \\
  Carnegie Mellon University \\
  {\tt \{chuntinz,xuezhem,junjieh,gneubig\}@cs.cmu.edu} \\}

\date{}

\begin{document}
\maketitle

\begin{abstract}
Despite impressive empirical successes of neural machine translation (NMT) on standard benchmarks, limited parallel data impedes the application of NMT models to many language pairs.
Data augmentation methods such as back-translation make it possible to use monolingual data to help alleviate these issues, but back-translation itself fails in extreme low-resource scenarios, especially for syntactically divergent languages.
In this paper, we propose a simple yet effective solution, whereby target-language sentences are re-ordered to match the order of the source and used as an additional source of training-time supervision. % Then these sentences, combined with a small amount of parallel data, are used to train an NMT model.
Experiments with simulated low-resource Japanese-to-English, and real low-resource Uyghur-to-English scenarios find significant improvements over other semi-supervised alternatives\footnote{https://github.com/violet-zct/pytorch-.reorder-nmt}
\end{abstract}

\section{Introduction}

While neural machine translation (NMT; \citet{bahdanau2014neural,vaswani2017attention}) now represents the state of the art in the majority of large-scale MT benchmarks \cite{bojar2017wmt}, it is highly dependent on the availability of copious parallel resources;
NMT under-performs previous phrase-based methods when the training data is small~\citep{koehn-knowles:2017:NMT}.
Unfortunately, million-sentence parallel corpora are often unavailable for many language pairs. 
Conversely, monolingual sentences, particularly in English, are often much easier to find, making semi-supervised approaches that can use monolingual data a desirable solution to this problem.%
\footnote{Unsupervised MT \cite{artetxe2018emnlp} has achieved success on simulated low-resource scenarios with related languages, but limited success on real low-resource settings and syntactically divergent language pairs \cite{neubig2018rapid,guzman2019two}. Hence we focus on semi-supervised methods in this paper.}

Semi-supervised approaches for NMT are often based on automatically creating pseudo-parallel sentences through methods such as back-translation \cite{
%munteanu:2004:HLTNAACL,
irvine2013combining,sennrich2015improving} or adding an auxiliary auto-encoding task on monolingual data \citep{cheng2016semi,he2016dual,currey2017copied}.
However, both methods have problems with low-resource and syntactically divergent language pairs.
Back translation assumes enough data to create a functional NMT system, an unrealistic requirement in low-resource scenarios, while auto-encoding target sentences by definition will not be able to learn source-target word reordering.

\begin{figure}[tb]
  \centering
  \includegraphics[scale=0.3]{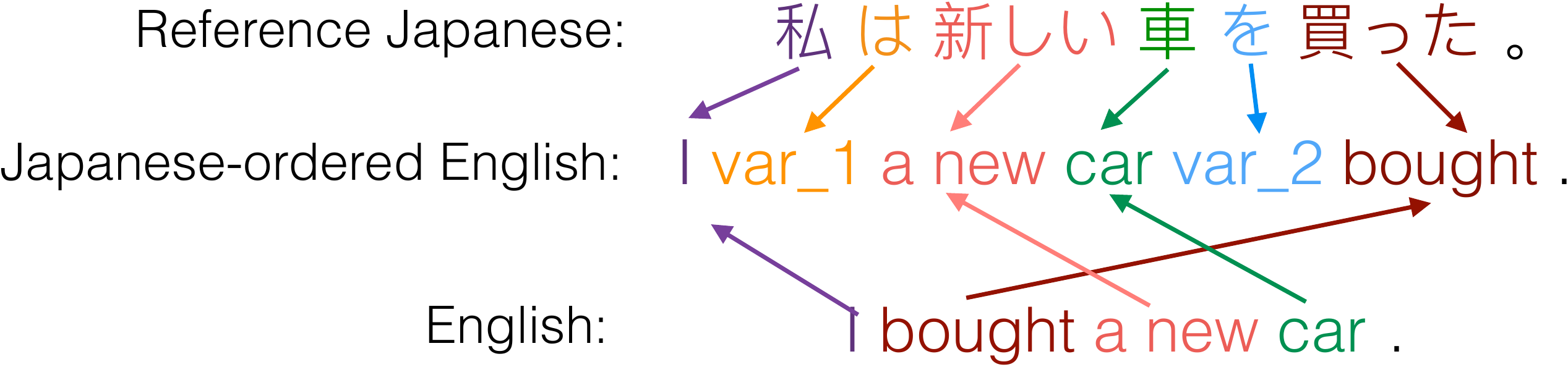}
  \caption{An English sentence re-ordered into Japanese order using the rule-based method of \newcite{isozaki2010head}, and its reference Japanese translation.} \label{fig:examples}
  \vspace{-3mm}
\end{figure}

This paper proposes a method to create pseudo-parallel sentences for NMT for language pairs with divergent syntactic structures. 
Prior to NMT, word reordering was a major challenge for statistical machine translation (SMT), and many techniques have emerged over the years to address this challenge~\citep{xia-mccord:2004:COLING,bisazza2016survey}.
Importantly, even simple heuristic reordering methods with a few hand-created rules have been shown to be highly effective in closing syntactic gaps  (\newcite{collins2005clause,isozaki2010head}; Fig.~\ref{fig:examples}).
Because these rules usually function solely in high-resourced languages such as English with high-quality syntactic analysis tools, a linguist with rudimentary knowledge of the structure of the target language can create them in short order using these tools. 

However, %the utility of explicit re-ordering has only been explored by few research work in NMT \cite{zhang2016exploiting,zhang2017incorporating}, 
similar pre-ordering methods have not proven useful in NMT \cite{du2017pre}, largely because high-resource scenarios NMT is much more effective at learning reordering than previous SMT methods were \cite{bentivogli-EtAl:2016:EMNLP2016}.
% We focus on a more difficult setting where the source language is distant from the target language which does not share the same alphabet and have different syntactic orders.
However, in low-resource scenarios it is less realistic to expect that NMT could learn this reordering from scratch on its own.

Here we ask ``how can we efficiently leverage the monolingual target data to improve the performance of the NMT system in low-resource, syntactically divergent language pairs?''
We tackle this problem via a simple two-step data augmentation method: (1) we first reorder monolingual target sentences to create \emph{source-ordered target sentences} as shown in Fig.~\ref{fig:examples}, (2) we then replace the words in the reordered sentences with source words using a bilingual dictionary, and add them as the source side of a pseudo-parallel corpus.
Experiments demonstrate the effectiveness of our approach on translation from Japanese and Uyghur to English, with a simple, linguistically motivated method of head finalization (HF; \citet{isozaki2010head}) as our reordering method. 
% Our approach achieves significant improvements over methods using only parallel data or other previous methods using pseudo-parallel corpora.

\section{The Proposed Method}
% First we give an overview of the proposed data-augmentation framework then give details of the major components.
% \vspace{-2mm}
\paragraph{Training Framework}
We assume that there are two types of available resources: a small parallel corpus $\mathcal{P}=\{(s, t)\}$ and a large monolingual target corpus $\mathcal{Q}$.
The goal of our method is to create a pseudo-parallel corpus $\hat{\mathcal{Q}}=\{(\hat{s}, t)\}$, where $\hat{s}$ is a pseudo-parallel sentence automatically created in two steps of (1) word reordering, and (2) word-by-word translation.
% There are three main steps in our proposed method: (i) ; (ii) . Then we can train an NMT model with pairs of the training examples from the large amount of pseudo parallel sentences $\mathcal{O}$ and the small amount of true parallel sentences $\mathcal{P}$. We call these ``reordered pairs'' and ``supervised pairs'' respectively. By augmenting the supervised data with a large number of reordered pairs, the NMT model is able to learn what information it can (e.g. a strong target-side language model or a reasonably high quality word reordering model), bootstrapping the training of more difficult phenomena (e.g. lexical mapping) that must be done on the supervised pairs.
\vspace{-2mm}
\paragraph{Word Reordering}
The first step reorders monolingual target sentences $t\in\mathcal{Q}$ into the source order $t_s$.
% Word reordering has been widely studied in 
% the context of
% statistical machine translation~\citep{bisazza2016survey}, and because our goal is to examine the utility of these methods in the context of data augmentation for NMT, we can 
Instead of devising an entirely new word-ordering method, we can simply rely on methods that have already been widely studied and proven useful in SMT~\citep{bisazza2016survey}.
Reordering can be done either using rules based on linguistic knowledge~\cite{isozaki2010head,collins2005clause} or learning from aligned parallel data~\cite{xia-mccord:2004:COLING,habash2007syntactic}, and in principle our pseudo-corpus creation paradigm is compatible with any of these methods.

Specifically, in this work we utilize rule-based methods, as our goal is to improve translation of low-resource languages, where large quantities of high-quality parallel data do not exist and we posit that current data-driven reordering methods are unlikely to function well.
Examples of rule-based methods include those to reorder English into German \cite{navratil2012comparison}, Arabic \cite{badr2009syntactic}, or Japanese \cite{isozaki2010head}.
In experiments we use \newcite{isozaki2010head}'s method of reordering SVO languages (e.g. English) into the order of SOV languages (e.g. Japanese) by simply (1) applying a syntactic parser to English \cite{tsuruoka2004enju}, (2) identifying the head constituent of each phrase and moving it to the end of the phrase, and (3) inserting special tokens after subjects and objects of predicates to mimic Japanese case markers.
% We apply this method to sentences $t\in\mathcal{Q}$ to create source-ordered target sentences $t_s$.
% \cz{I have explained how we do reordering in the section of experimental setups (parsing and restructuring).}
% \gn{I added one more sentnece to explain here, because I think it's necessary}
\vspace{-2mm}
\paragraph{Word-by-word Translation}
% Creating a bilingual word dictionary $\mathcal{D}$ manually or learning it from available small amounts of supervised data or in an unsupervised fashion; (iii) Creating a pseudo parallel corpora with (\emph{pseudo-source, target}) sentence pairs denoted $\mathcal{O} = \{(\hat{s}, t)\}$ where $t\in\mathcal{Q}$ and $\hat{s}$ is translated word-by-word from the source-ordered target sentence $t_s$ via the bilingual dictionary $\mathcal{D}$
To generate data for training MT models, we next perform word-by-word translation of $t_s$ into pseudo-source sentence $\hat{s}$ using a bilingual dictionary \cite{xie2018neural}.%
\footnote{We also performed extensive preliminary experiments that learned bilingual word embeddings a-priori and froze them when training the NMT model, or continued to align the bilingual word embedding space during NMT training, but the word-by-word translation approach worked best.}
% There are three ways of utilizing the source-ordered target sentences $t_s$: (i) learn bilingual word embeddings and freeze them when training the NMT model; (ii) continue aligning the bilingual word embedding space during training the NMT model and (iii)  In our experiments, we find that the third approach works best while the other two options make learning difficult and yield worse results. We create the pesudo-source sentences $\hat{s}$ by replacing words in $t_s$ with the source language words using bilingual lexicon induction and available bilingual dictionaries. After such noisy translation, we obtain a pseudo parallel corpora $\mathcal{O}$ and use that to augment the limited parallel data $\mathcal{P}$.
There are many ways we can obtain this dictionary: even for many low-resource languages with a paucity of bilingual text, we can obtain manually-curated lexicons with reasonable coverage, or run unsupervised word alignment on whatever parallel data we have available.
In addition, we can induce word translations for more words in target language using methods for bilingual lexicon induction over pre-trained word embeddings (e.g.~\citet{grave2018learning}).
% A shared bilingual embedding space can be learned either using supervised methods that rely on a seed lexicon such as the ones described above \cite{jawanpuria2018learning} or unsupervised methods that have no such resource requirements \cite{artetxe2018robust}
% We then adopt the cross-domain similarity local scaling (CSLS) metric \cite{lample2018word} to induce the nearest neighbor of a word which is added to our augmented lexicon. 

\section{Experiments}
We evaluate our method on two language pairs: Japanese-to-English (\texttt{ja-en}) and Uyghur-to-English (\texttt{ug-en}).
Japanese and Uyghur are phylogenetically distant languages, but they share similar SOV syntactic structure, which is greatly divergent from
English SVO structure.
% We first give the common experimental setups shared between the two language pairs and then describe the tailored settings respectively.
\vspace{-2mm}
\subsection{Experimental Setup}
For both language pairs, we use an attention-based encoder-decoder NMT model with a one-layer bidirectional LSTM as the encoder and one-layer uni-directional LSTM as the decoder.%
\footnote{We experimented with small Transformers \cite{vaswani2017attention} but they under-performed LSTM-based models.}
Embeddings and LSTM states were set to 300 and 256 dimensions respectively. Target word embeddings are shared with the softmax weight matrix in the decoder.
As noted above, we use HF \cite{isozaki2010head} as our re-ordering rule.
HF was designed for transforming English into Japanese order, but we use it \emph{as-is} for the Uyghur-English pair as well to demonstrate that simple, linguistically motivated rules can generalize across pairs with similar syntax with little or no modification.
Further details regarding the experimental settings are in the supplementary material.
\vspace{-2mm}
\paragraph{Simulated Japanese to English Experiments}
We first evaluate on a simulated low-resource \texttt{ja-en} translation task using the ASPEC dataset~\cite{NAKAZAWA16.621}. We randomly select 400k \texttt{ja-en} parallel sentence pairs to use as our full training data.
We then randomly sub-sample low-resource datasets of 3k, 6k, 10k, and 20k parallel sentences, and use the remainder of the 400k English sentences as monolingual data. We duplicate the number of parallel sentences by 5 times in the training data augmented with the reordered pairs.
For settings with supervised parallel sentences of 3k, 6k, 10k and 20k, we set the maximum vocabulary size of both Japanese and English to be 10k, 10k, 15k and 20k respectively.

To automatically learn a high-precision dictionary on the small amount of parallel data we have available for training, we use GIZA++~\cite{och03:asc} to learn alignments in both directions then take the intersection of alignments.
% We then learn the bilingual word embeddings with the unsupervised density matching method \cite{zhou19naacl}.
% We then learn the bilingual word embeddings with the unsupervised density matching method \footnote{\url{https://github.com/violet-zct/flow-bme}}.
We then learn the bilingual word embeddings with DeMa-BWE \cite{zhou19naacl}, an unsupervised method that has shown strong results on syntactically divergent language pairs. We give the more reliable alignments extracted from GIZA++ high priority by querying the alignment dictionary first, then follow by querying the embedding-induced dictionary. When an English word is not within any vocabulary, we output the English word as-is into the pseudo-source sentence.

\begin{table}[t]
\centering
\resizebox{\columnwidth}{!}{%
\begin{tabular}{lccrrrrrr}
\toprule
Model & 3k & 6k & 10k & 20k & 400k & ug\\
\midrule
% unsup$^1$ & -- & -- & -- & -- & 0.0 \\
% unsup$^2$ & -- & -- & -- & -- & 0.6 \\
sup & 2.17 & 7.86 & 11.67 & 15.98 & 26.56 & 0.58\\
sup-SMT & 6.36 & 8.70 & 10.68 & 12.11 & 18.62 & 1.46 \\
back & 2.27 & 5.40 & 13.50 & 16.05 & -- & 0.42 \\
back-SMT & 8.46 & 10.61 & 12.05 & 13.68 & -- & 1.37 \\
No-Reorder &  6.46 & 9.73 & 12.57 & 15.56 & -- & 3.24\\
\midrule
Reorder &  \textbf{9.94} & \textbf{12.42} & \textbf{14.98} & \textbf{17.58} & -- & \textbf{4.17}\\
\bottomrule
\end{tabular}
}
\caption{BLEU of our approach (Reorder) with different amount of parallel sentences of \texttt{ja-en} and \texttt{ug-en} translation. Baselines are supervised learning from NMT and SMT (sup and sup-SMT), supervised learning with back translation from NMT and SMT (back and back-SMT) and data augmentation with translated original English sentences (No-Reorder).}
\label{tab:res:mt}
\vspace{-2mm}
\end{table}

\begin{table*}
\centering
\small
\begin{tabular}{p{1.5cm}p{13.5cm}}
\toprule
\textbf{source} & \pbox{13cm}{\begin{CJK}{UTF8}{min} しかし ， 回転 速度 が 大き すぎ る と ， 逆向き の 変形 が 生じ る \end{CJK}}\\
\textbf{reference} &  the too high rotation speed produces the reverse deformation \\
\textbf{supervised} &  however , the deformation of <unk> and the deformation of <unk> is caused by the dc rate \\
\textbf{ours} & however , the deformation of <unk> is generated when the rotation rate is large\\
\midrule
\textbf{source} & 
\setuighur
\substituteNumbers{%
\begin{arabtext}
2 - ayning 12 - küni , Ürümchi vaqti 3:29 Minutta 3.3 Bal yer tevrigen , yer tevresh chongqurliqi 8000 Mëtir .
\end{arabtext}
}
% \LR{2 -} \begin{RLtext} Ayning \end{RLtext} \LR{12 -}  k\"{u}ni %, '\"{u}r\"{u}mqi va\kappa ti 3:29 minutta 3.3 bal y\inve r t\inve vrig\inv n , y\inv r t\inve vr\inve x qong\kappa urli\kappa i metir .
% 2 - ئاينىڭ 12 - كۈنى , ئۈرۈمچى ۋاقتى 3:29مىنۇتتا 3.3 بال يەر تەۋرىگەن , يەر تەۋرەش چوڭقۇرلىقى 8000 مېتىر .
\vspace{-0.2cm}
\\
\textbf{reference} & a 3.3 magnitude earthquake with the depth of 8000 meters hit feb 12 at 3:29 urumqi time \\
\textbf{supervised} & 2 , on february 12 - 12 of darkness , urumqi time hit a 3.3 earthquake , the earthquake hit . \\
\textbf{ours} & 2 - on february 12 - on , urumqi time 3:29minus 3.3 magnitude earthquake hit , the earthquake under depth of 8000 meters .\\
\bottomrule
\end{tabular}
\caption{Translation examples on \texttt{ja-en} (reorder with 6000 supervised pairs) and \texttt{ug-en} (reorder) from our model and the supervised counterpart.}
\label{tab:example}
\vspace{-3mm}
\end{table*}

% We then make the intersection of the word alignments of two directions.
\vspace{-2mm}
\paragraph{Real Uyghur to English Experiments} We also consider the harder case of Uyghur, a truly low-resource language. We create test and validation sets using the test data from the DARPA LORELEI corpus \cite{Christianson:2018:ODL:3237031.3237070} which contains 2,275 sentence pairs (after filtering out noisy ones) related to incidents that happened in the Uyghur area. We hold out 300 pairs as the validation data and use the rest as the test set. The LORELEI language pack also contains the bilingual lexicons between Uyghur and English, and thousands of in-domain English sentences. We also use a large monolingual English corpus containing sentences related to various incidents occurring all over the world collected from ReliefWeb.\footnote{\url{https://reliefweb.int}} To sub-select a relevant subset of this corpus, we use the cross-entropy filtering \cite{moore2010intelligent} to select 400k that are most like the in-domain English data.

For parallel data, like many low-resource languages, we only have access to data from the Bible\footnote{\url{https://bible.is}} and Wikipedia language links (the total number of parallel Uyghur-English Wikipedia titles is 3,088), but no other in-domain parallel data. 
% We select 30k most relevant (incident-related) sentence pairs with cross-entropy filtering from the Bible data which we find perform better than using the full Bible data set of Uyghur.
We run GIZA++ on this parallel data to obtain an alignment dictionary. We learn the bilingual word embeddings via the supervised Geometric approach \cite{jawanpuria2018learning} on FastText \cite{grave2018learning} pre-trained Uyghur and English monolingual embeddings.

% Then similarly to \texttt{ja-en} translation, we re-order the sentences and translate the in-domain reordered English sentences via the bilingual lexicons. \cz{I have already simplified the steps of how to create in-domain resources for Uyghur. Do we need to add more details of it like the statistics of training data, dictionary, embeddings, Wikipedia links, etc.} 
% \gn{I was also thinking that this lacked detail, maybe we could add it to the appendix.}

% \begin{figure}[tb]
%   \centering
%   \includegraphics[width=1.0 \columnwidth]{figs/ja_ribes.pdf}
%   \vspace{-6mm}
%   \caption{Comparison of RIBES score on Ja-En translation with different amounts of supervised data.}
%   \label{fig:ribes}
%   \vspace{-4mm}
% \end{figure}

\begin{figure}[tb]
  \centering
  \includegraphics[width=1.0 \columnwidth]{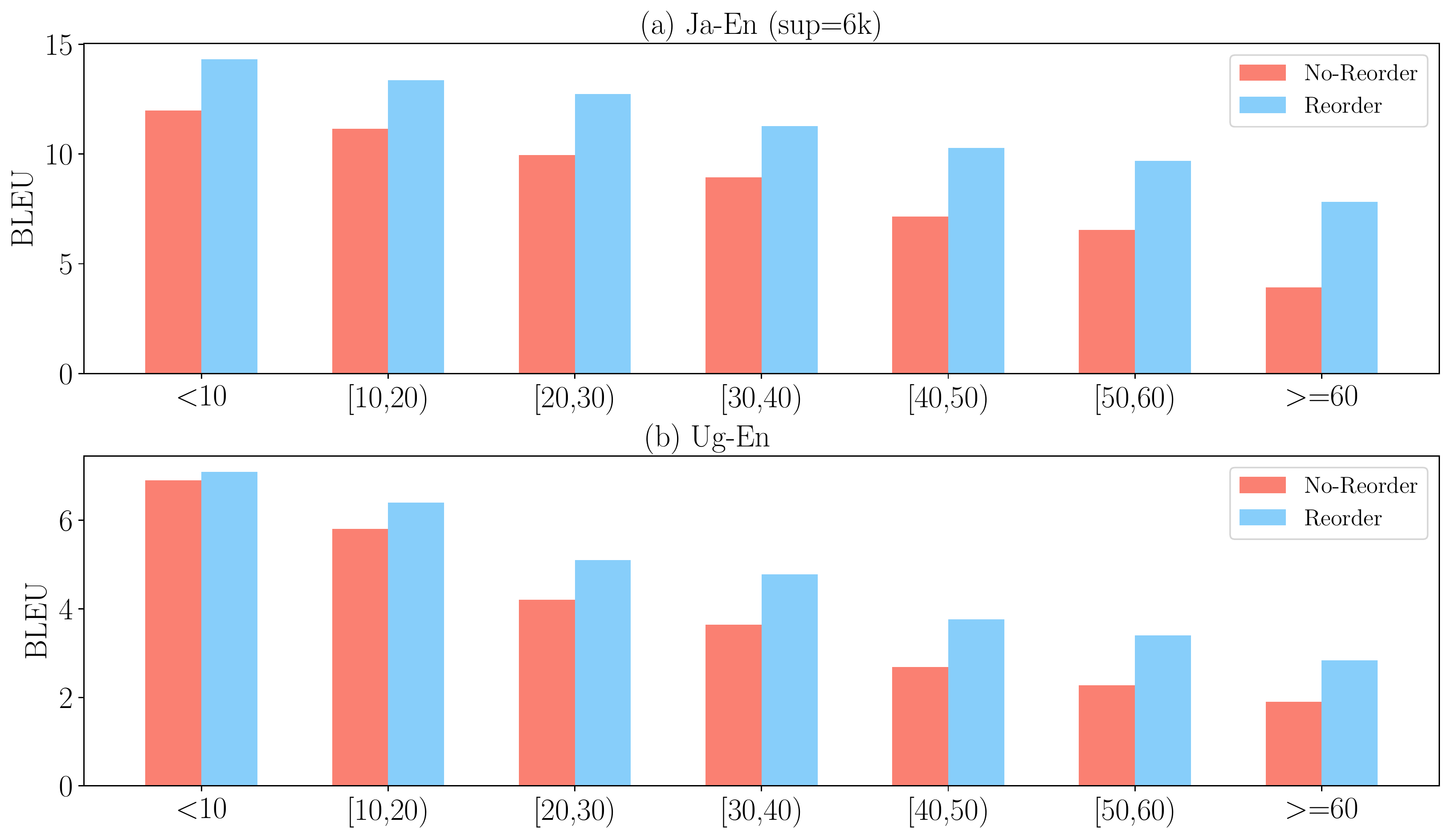}
  \vspace{-2mm}
  \caption{Comparison of BLEU score w.r.t different sentence lengths.}
%   The Ja-En results are from the model with 6k supervised data.}
  \label{fig:length}
  \vspace{-2mm}
\end{figure}

\subsection{Results and Comparison}
\paragraph{Baselines}
\label{sec:baseline}
In Tab.~\ref{tab:res:mt}, we compare our models with baselines including regular supervised training (sup) and back-translation~\citep{sennrich2015improving} (back).%
\footnote{We also trained unsupervised NMT~\cite{artetxe2017unsupervised} on 400k Japanese and English sentences from the ASPEC corpus or monolingual Uyghur and English corpus provided in the LORELEI data package, and it achieved BLEU of 0.6 and 0.0 respectively, corroborating previous results noting the difficulty of unsupervised NMT for low-resource scenarios \cite{neubig2018rapid,guzman2019two}}
To demonstrate the effectiveness of the reordering, we also compare our method against a copy-based data-augmentation method (No-reorder) where the original English sentences $t\in \mathcal{Q}$ rather than the reordered ones $t_s$ are translated via the bilingual lexicon.%
\footnote{This is similar to \newcite{currey2017copied} who simply copy target-language sentences to the source. But in our experiments, the source and target languages do not share the alphabet and thus we found translation necessary. This also increases consistency with our ``Reorder'' experiments.}
For each of the above settings, we also experimented with a phrased-based statistical machine translation (SMT) system \cite{dyer2010cdec}. In Tab. \ref{tab:res:mt}, we only show the results with supervised data and back-translation for SMT, since we observed that the data augmentation method performs poorly with SMT (complete results are presented in the Appendix).

% \vspace{-2mm}
\paragraph{Main Results}
In Tab. \ref{tab:res:mt}, we observe consistent improvements on both \texttt{ja-en} and \texttt{ug-en} translation tasks against other baseline methods.
First, comparing our results with the NMT models trained using the same amount of parallel data, our word reordering-based semi-supervised models consistently outperform standard NMT models by a large margin. In the case that we have no access to in-domain parallel data at all, our method can still achieve some success in \texttt{ug-en} translation. 
Second, comparing our Reorder method with the No-Reorder one, reordering English sentences into the source language order consistently brings large performance gains, which demonstrates the importance of reordering.
These results are notable given previous reports that explicit reordering is not beneficial for NMT \cite{du2017pre}.
Third, for \texttt{ja-en} translation, when gradually decreasing the amount of parallel data, the improvements of our model over the supervised NMT models become more significant, demonstrating the effectiveness of our approach in low-resource settings. 
Fourth, back-translation is not very beneficial or even harmful, likely because the back-translation system trained on limited supervised data can not provide high-quality translations to train the model.
Finally, we also notice that although SMT performs better than NMT with less supervised training data (3k, 6k supervised data and Uyghur), the performance gain is not as remarkable as NMT when the amount of supervised data increases. Moreover, in the case of less supervised data, our data augmentation method with reordering still outperforms SMT.

\begin{figure}[hb]
  \centering
  \includegraphics[width=0.9 \columnwidth]{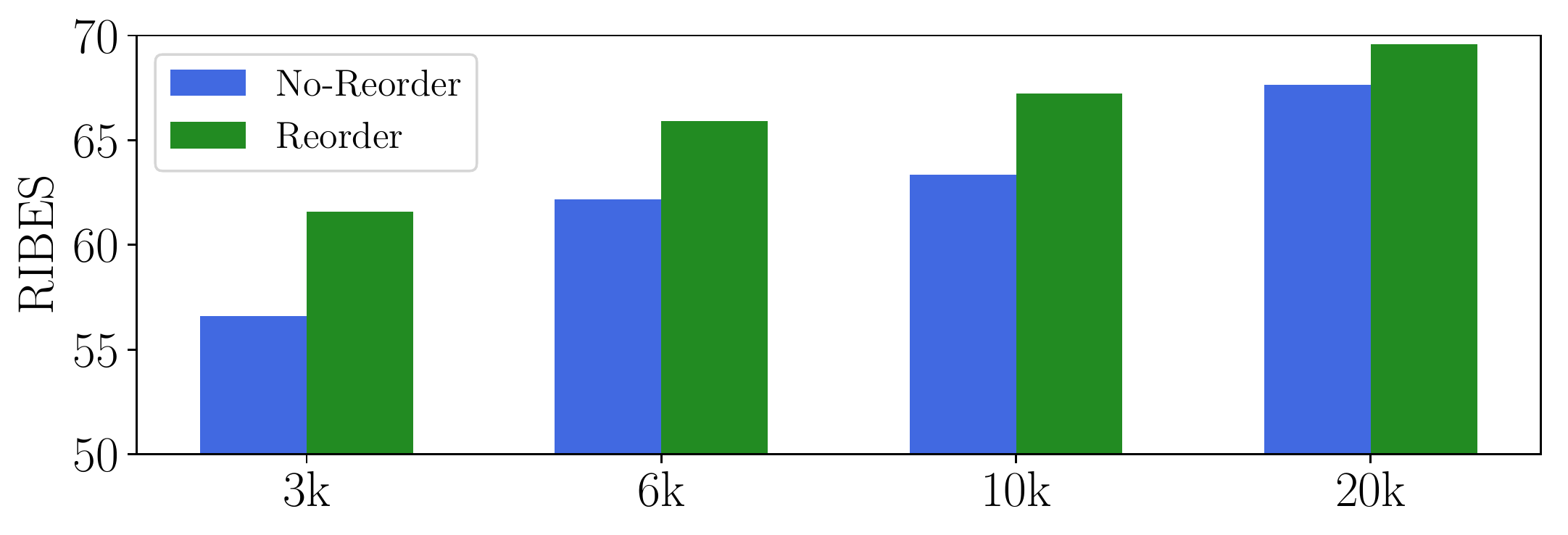}
%   \vspace{-6mm}
  \caption{Comparison of RIBES score on Ja-En translation with different amounts of supervised data.}
  \label{fig:ribes}
%   \vspace{-4mm}
\end{figure}

We give two examples of the translation outputs of our model and a supervised NMT model for \texttt{ug-en} and \texttt{ja-en} (trained with 6k supervised pairs) in Tab.~\ref{tab:example}. In the first example from \texttt{ja-en}, our model is able to output terminology such as ``rotation rate'' thanks to the enlarged vocabulary while the supervised model can not. In the example from \texttt{ug-en}, our model can produce a more fluent sentence with better information coverage.

\paragraph{Analysis}
To investigate the effects of reordering, we compare our method with ``No-Reorder" described in \ref{sec:baseline}. 
First, we bucket the test data by sentence length and compute the BLEU score accordingly. We present the comparison results in Fig.~\ref{fig:length}, from which we observe that ``Reorder" outperforms ``No-Reorder" consistently under different sentence length buckets, and the improvement is larger when the sentence length is longer.

Second, we also evaluate the model outputs on the test data with RIBES \cite{isozaki2010automatic}
which is an automatic evaluation metric of translation quality designed for distant languages that is especially sensitive to word order. From Fig.~\ref{fig:ribes}, we can see that ``Reorder" consistently outperforms ``No-Reorder" on \texttt{ja-en} translation especially when the amount of supervised data decreases. This suggests that with reordered pairs as the augmented training data, the model is able to output more syntactically correct sentences.
% \begin{table}[t!]
% \centering
% \resizebox{\columnwidth}{!}{
% \begin{tabular}{ccccc}
% \toprule
% unsup & back & sup & No-Reorder & Reorder\\
% \midrule
% 0.0 & 0.42 & 0.58 & 3.67 & \textbf{4.71} \\
% \bottomrule
% \end{tabular}}
% \caption{BLEU of our approach (Reord) against baseline methods on \texttt{ug-en} test set.}
% \label{tab:uyg}
% \vspace{-3mm}
% \end{table}

% \section*{Acknowledgments}
\section{Conclusion}
This paper proposed a simple yet effective semi-supervised learning framework for low-resource machine translation that artificially creates source-ordered target sentences for data-augmentation. 
%To ensure aligned bilingual word embeddings, we learn an in-domain bilingual dictionary from a small amount of parallel data and optionally add an L2 loss to make the bilingual embedding space stay aligned during training.
Experimental results on \texttt{ja-en} and \texttt{ug-en} translations show that our approach achieves significant improvements over baseline systems, demonstrating the effectiveness of the proposed approach on divergent language pairs.

\section*{Acknowledgments}
This work is sponsored by Defense Advanced Research Projects Agency Information Innovation Office (I2O), Program: Low Resource Languages for Emergent Incidents (LORELEI), issued by DARPA/I2O under Contract No. HR0011-15-C-0114. 
The authors would like to thank Shruti Rijhwani and Hiroaki Hayashi for their help when preparing the data sets.

\newpage
\bibliography{emnlp-ijcnlp-2019}
\bibliographystyle{acl_natbib}

\appendix
\label{sec:appendix}
\onecolumn
\begin{center}
\Large
\textbf{Handling Syntactic Divergence in \\
Low-resource Neural Machine Translation}

Supplementary Materials
\end{center}

\section{Results}
We present the full results in Tab.~\ref{tab:res:mt:full}, from which we can see that as the amount of supervised data increases, the performance gain of SMT is not as much as the NMT model. For SMT, reordering has much better performance than no-reorder, but still lags behind the supervised counterpart.
\begin{table*}[h]
\centering
\resizebox{\columnwidth}{!}
{%
\begin{tabular}{lcccccccccccc}
\toprule
 & \multicolumn{2}{c}{3k} & \multicolumn{2}{c}{6k} & \multicolumn{2}{c}{10k} & \multicolumn{2}{c}{20k} & \multicolumn{2}{c}{400k} & \multicolumn{2}{c}{ug}\\
Model & NMT & SMT & NMT & SMT & NMT & SMT & NMT & SMT & NMT & SMT & NMT & SMT \\
\midrule
% unsup$^1$ & -- & -- & -- & -- & 0.0 \\
% unsup$^2$ & -- & -- & -- & -- & 0.6 \\
sup & 2.17 & 6.36 & 7.86 & 8.70 & 11.67 & 10.68 & 15.98 & 12.11 & 26.56 & 18.62 & 0.58 & 1.46 \\
back & 2.27 & 8.46 & 5.40 & 10.61 & 13.50 & 12.05 & 16.05 & 13.68 & -- & -- & 0.42 & 1.37\\
No-Reorder &  6.46 & 3.08 & 9.73 & 5.24 & 12.57 & 6.72 & 15.56  & 8.96 & -- & -- & 3.24 & 1.67 \\
\midrule
Reorder &  \textbf{9.94} & 6.23 & \textbf{12.42} & 8.14 & \textbf{14.98} & 9.22  & \textbf{17.58} & 11.21 & -- & -- & \textbf{4.17} & 1.07 \\
\bottomrule
\end{tabular}
}
\caption{BLEU of our approach (Reorder) with different amount of parallel sentences of \texttt{ja-en} and \texttt{ug-en} translation. Baselines are supervised learning (sup), supervised learning with back translation (back) and data augmentation with translated original English sentences (No-Reorder).}
\label{tab:res:mt:full}
\vspace{-2mm}
\end{table*}

% \section{Other Experimental Setup on Japanese-English Translation}

% \section{Uyghur-English Resource Details}
% \paragraph{Bible Data} We select 30k most relevant sentence pairs based on Moore-Lewis method \cite{moore2010intelligent} which we find perform better than using the full Bible data set of Uyghur.
% \paragraph{Bilingual Lexicon} We cleaned up and combined the dictionaries contained in the DARPA LORELEI corpus. The total number of items in the collected dictionaries is around 17k.
% \paragraph{Wikipedia Links} The total number of parallel Uyghur-English Wikipedia titles is 3,088.
% \section{Analysis on RIBES Score}

\end{document}